\documentclass{article}

\usepackage{PRIMEarxiv}

\usepackage[utf8]{inputenc} % allow utf-8 input
\usepackage[T1]{fontenc}    % use 8-bit T1 fonts
\usepackage{hyperref}       % hyperlinks
\usepackage{url}            % simple URL typesetting
\usepackage{booktabs}       % professional-quality tables
\usepackage{amsfonts}       % blackboard math symbols
\usepackage{nicefrac}       % compact symbols for 1/2, etc.
\usepackage{microtype}      % microtypography
\usepackage{lipsum}
\usepackage{fancyhdr}       % header
\usepackage{graphicx}       % graphics
\graphicspath{{media/}}     % organize your images and other figures under media/ folder
\usepackage{amsmath}
\usepackage{amssymb}
\usepackage{bm}
\usepackage{algorithm, algorithmic}
\usepackage{cite}

\title{A Learning System for Motion Planning of Free-Float Dual-Arm Space Manipulator
towards Non-Cooperative Object
}

\author{
  Shengjie Wang \\
  Department of Automation \\
  Tsinghua University \\
  Beijing\\
  \texttt{wangsj19@mails.tsinghua.edu.cn} \\
   \And
  Yuxue Cao \\
  Beijing Institute of Control Engineering \\
  Beijing\\
  \texttt{sg18801138270@163.com} \\
   \And
  Xiang Zheng \\
  Department of Computer Science \\
  City University of Hong Kong \\
  Hong Kong\\
  \texttt{xzheng235-c@my.cityu.edu.hk} \\
   \And
  Tao Zhang$^{*}$ \\
  Department of Automation \\
  Tsinghua University \\
  Beijing\\
  \texttt{taozhang@tsinghua.edu.cn} \\
\thanks{\textit{\underline{Corresponding author}}: 
\textbf{Tao Zhang}(taozhang@tsinghua.edu.cn)}
}

\begin{document}
\maketitle

\begin{abstract}
Recent years have seen the emergence of non-cooperative objects in space, like failed satellites and space junk. These objects are usually operated or collected by free-float dual-arm space manipulators. Thanks to eliminating the difficulties of modeling and manual parameter-tuning, reinforcement learning (RL) methods have shown a more promising sign in the trajectory planning of space manipulators. Although previous studies demonstrate their effectiveness, they cannot be applied in tracking dynamic targets with unknown rotation (non-cooperative objects). In this paper, we proposed a learning system for motion planning of free-float dual-arm space manipulator (FFDASM) towards non-cooperative objects. Specifically, our method consists of two modules. Module I realizes the multi-target trajectory planning for two end-effectors within a large target space. Next, Module II takes as input the point clouds of the non-cooperative object to estimate the motional property, and then can predict the position of target points on an non-cooperative object. We leveraged the combination of Module I and Module II to track target points on a spinning object with unknown regularity successfully. Furthermore, the experiments also demonstrate the scalability and generalization of our learning system.

\end{abstract}

% keywords can be removed

\section{Introduction}

With the continuous increase of non-cooperative space objects like faulty satellites and space junk, space manipulators become indispensable in many space maintenance tasks.  Free-float dual-arm space manipulator (FFDASM) is a common configure for space manipulators, of which the free-float advantage helps to save fuel consumed in adjusting the pose and the dual-arm structure enlarges the workspace efficiently \cite{oda1996ets}. However, the characteristic of FFDASM has some negative impacts on model-based algorithms in trajectory planning tasks. Due to the coupling effects between the base and two manipulators, there are many time-varying and dynamic parameters in the Jacobian matrix,  which are hard to be identified  \cite{yoshida2001zero,wang2009passivity}. Furthermore, trajectory planning aiming for a single target should take the joint trajectories into account, introducing more biases in controllers \cite{liu2016trajectory,yoshida2006capture}. What's more, it will take a long time to manually adjust parameters for experts in order to achieve the full potential of these model-based controllers  \cite{yan2021trajectory,huang2007adaptive}.

\begin{figure}[t]
  \centering
  \includegraphics[width=\hsize]{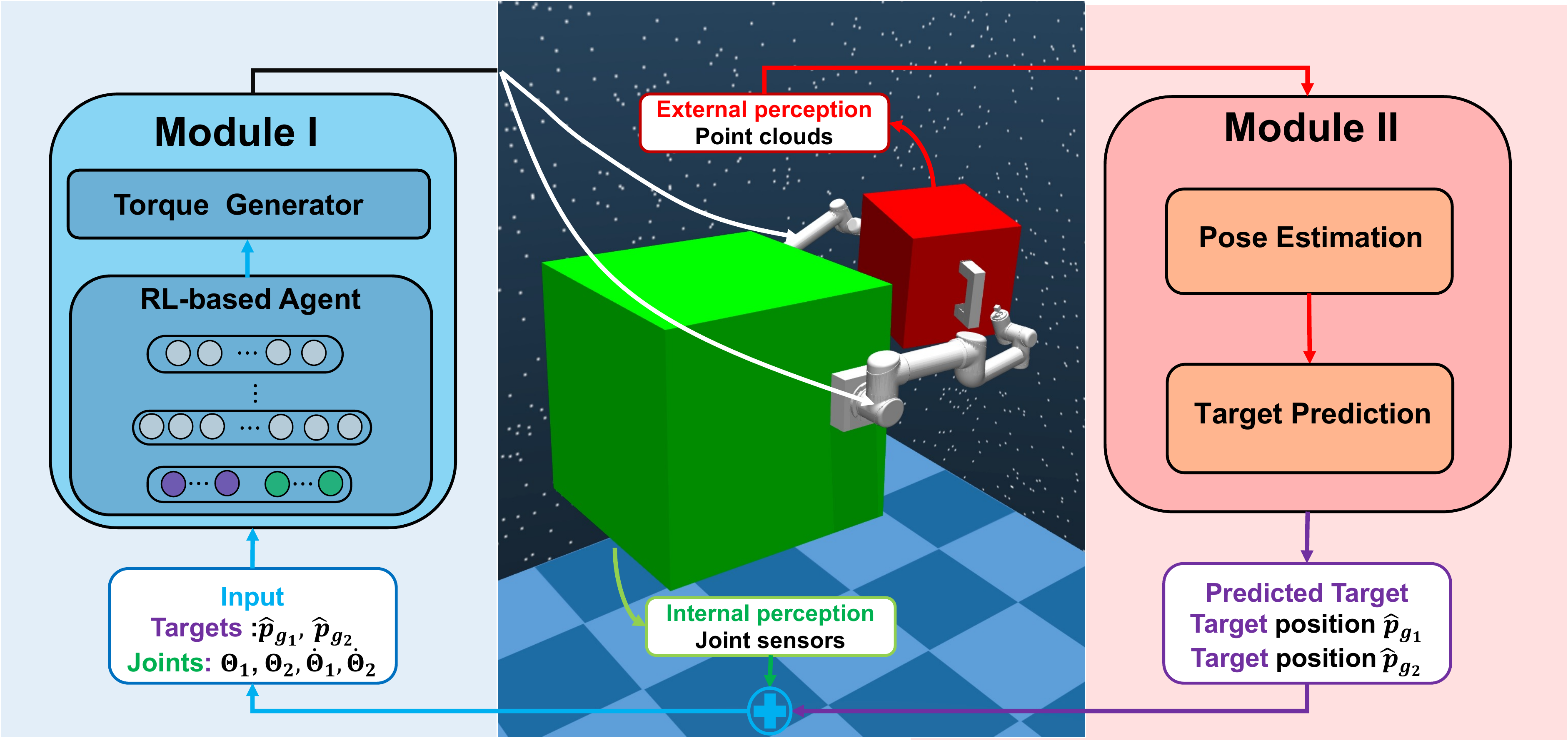}
  \caption{Framework of our two-module learning system for motion planning of free-float dual-arm space manipulator. Module I estimates the pose of spinning object, and predict the target points. Module II receives the prediction, and then allows the space manipulator to track the targets by a RL-based agent.}
  \label{fig1}
% \vspace{-5ex}
\end{figure}

To eliminate the mentioned problems above, model-free methods attracts more attentions from researchers. Model-free methods doesn't make use of approximate kinematic and dynamic models of space manipulators. In the meantime, the controller is built by a polynomial function, bessel curve function, or neural network with learnable parameters \cite{WANG20181525}. Intuitively, these functions only take some part of feedback from sensors as input, without concern of modeling. To reach the target, we can view the trajectory planning as an optimization problem naturally. Then, the optimization method updates the parameters of functions at each iteration. After the convergence, the optimal controller is obtained to perform the task. To recap, the essential components are function of controller and optimization algorithm. Trajectory planning for fixed start and end is able to be solved by the combination of polynomial controllers and some non-linear optimization methods, such as PSO and GA methods \cite{wang2015trajectory,chen2017path,WANG20181525}. In comparison, reinforcement learning methods lead to a dramatic improvement in robustness and generalization, especially for the field of robotics \cite{lee2020learning,chen2022system,yu2020meta,zhong2021collision,yamada2020motion,sang2020SelfConfiguring}. Interestingly, concurrent studies have also revealed the promising results on the intersection of reinforcement learning and space robots.

However, to realize a real-world planning controller based on reinforcement learning, we need to overcome a number of major challenges. The first problem is that most previous studies pay more attention to the single fixed target during the training, which means it is unable to handle the sim-to-real gap, e.g., static target in another position or dynamic one with unknown kinematics \cite{yan2018control,2020Reinforcement,li2021constrained}. Therefore, multi-target trajectory planning remains an existing task to be done. Nevertheless, standard algorithms encounter sample inefficiency towards multiple targets. Furthermore, although the previous study realizes the task in which two end-effectors track moving targets in the learning phase, it still needs to re-train policies for each rotating speed \cite{li2021constrained} when deployed in the test environment. The underlying reason is that the lack of perception module restricts the practical application in different environments. Thus, we need to provide enough measure information for the real-world practice, for example a non-cooperative object's angular velocity. To sum up, there remains an open question on developing a motion planning system for moving target points on a non-cooperative object.

To address the above problems, the main contribution of this paper is a learning system for motion planning of the free-float dual-arm space manipulator towards non-cooperative objects. The essential function of the system is to realize the trajectory planning for target points on the surface of a spinning object by Module I and Module II, which is shown in Fig. \ref{fig1}. We further made the following contributions:
\begin{itemize}
\item We proposed the Constrained Hindsight Experience Replay (C-HER) algorithm for multiple targets within a large space, improving the sample efficiency of standard actor-critic framework. 
% 后面需要补一个算法流程图
% \item We introduced the practical constraints into the objective function, and evaluated the performance of Penalty method and Larangian method on the balance between reward and cost. 
\item We developed the target's pose prediction module utilizing the point clouds of objects in space, which firstly estimates the rotating axis and angular speed of object, and then predicts the pose of target points.
\item From the perspective of space robotics, this work deploys learning-based perception and planning modules in tracking target points on a non-cooperative object successfully.
\end{itemize}
The combination of representation learning and reinforcement learning makes our system possible to track a non-cooperative object without the prior knowledge of the spinning speed, so as to easily generalize to diverse environments in the future.

\section{Related Work}

In contrast to model-based methods, model-free motion planning methods avoid potential modeling errors of the complex system dynamics. Also, they didn't need experts to manually adjust the parameters of modular controllers. The development of model-free methods can be roughly divided into two periods. In the early time, researchers applied various searching approaches to solving the optimization problem of trajectory planning. For instance, population-based methods including Differential Evolution (DE) \cite{WANG20181525}, Particle Swarm Optimization (PSO) \cite{wang2015trajectory} and Genetic Algorithm (GA) \cite{chen2017path} can be leveraged to obtain the optimal joint trajectory parameterized by polynomial functions, B-Spline or Bézier curve. The results demonstrate the effectiveness when the start and end are fixed, but the uncertainty of model and disturbances have a huge impact on performance.

Hereafter, due to the significant generalization, deep reinforcement learning methods have become popular in the field of robotics, especially for the exploration of quadruped robots in unstructured environments \cite{lee2020learning,tsounis2020deepgait}, in-hand object reorientation of a multi-finger robotic hand \cite{andrychowicz2020learning,chen2022system}, and multi-robot path planning \cite{long2018optimally,prianto2020path,ha2020learning}.  For the free-floating space manipulator, Yan \emph{et al.} proposed a trajectory planning method based on Soft Q-learning for a 3-DoF free-floating space robot \cite{yan2018control}. To reach multiple targets within a large space, Wang \emph{et al.} developed an improved version of Proximal Policy Optimization (PPO) for a 6-DoF space robot \cite{Wang20221multi}. However, in our experiments, the method can not work well in a 12-DoF dual-arm environment. The subsequent work mainly focus on the position and attitude decoupling control for the setting of a single robotic arm \cite{wang2022collision}. Recently, there are some preliminary applications for free-float dual-arm space manipulator. Wu \emph{et al.} applied Deep Deterministic Policy Gradient (DDPG) algorithm with the dense reward function, to realize the task for 8-DoF FFDASM \cite{2020Reinforcement}. Then,  Li \emph{et al.} considered the final velocity of end-effector into reward function to improve the previous method, but the results aren't evaluated by comparison \cite{li2021constrained}. In their later work, DDPG was combined with APF to improve the convergence of the algorithm, using potential energy difference for the reward calculation\cite{li2022constrained}. Furthermore, instead of introducing the constraints in the reward function, advanced methods usually take the combination of constraints and objectives into account during optimization, like Penalty method and Larangian method. Interestingly, their method trained a policy that can help the end-effectors reach the targets on the object, the spinning velocity of which is known constant. In this work, we solved trajectory planning for the spinning objects under unknown kinematic law, because we provided practical approaches to estimate the rotation of the object, and predict the position of targets. 

\section{Problem Statement}
\subsection{Modeling of Free-Float Dual-Arm Space Manipulator}

The core components of FFDASM are a base satellite and two robotic manipulators, Arm-1 and Arm-2 respectively. Two manipulators are fixed rigidly on the base, and the gravity is ignored in the simulation. Concretely, we chose two 6-DoF UR-5 robots as the manipulators, and the cubic structure as the base. Due to its slight impact, we removed the solar array for simplicity. Furthermore, the default kinematic and dynamic parameters are similar to the real space robot.

We first defined some notions. Arm-1 and Arm-2 represent two mission arms, the end-effectors of which need to reach the targets in the work space individually. $p_{e_i}$ and $v_{e_i}$ denote the position and velocity of the end-effector of Arm-$i$. $\theta^j_i$  is the angle of $j$-th joint of Arm-$i$, ranging from $[-\pi,\pi]$, $\Theta_i = [\theta^1_i,\theta^2_i,...,\theta^6_i]$. In this case, $\dot{\theta^j_i}$  naturally represents the angular velocity, and $\tau^j_i$ means the torque of the joint. Moreover, the pose vector of the end-effector and base are expressed as ${r}_{e_i}$and $r_b$.

Based on kinematic equation, the velocity of the end-effector of Arm-i can be formulated as:
\begin{equation}
  \label{eq2}
\dot {r}_{e_i} =J_{b}\dot {r_{b}}+J^i_{r} \dot{\Theta}_{i}
\end{equation}
Where $J_b$ is the Jacobian matrix with regard to the base, and $J^i_r$ is the Jacobian matrix of Arm-$i$.

Assuming the initial linear and angular momentum of system is zero, we can derive the momentum conservation equation:
\begin{equation}
  \label{eq3}
H_{b}\dot {r_{b}}+H^1_{r} \dot{\Theta}_{1}+ H^2_{r} \dot{\Theta}_{2}=0
\end{equation}
where $H_b$, $H^1_r$ and $H^2_r$ are the coupling inertia matrices of base, Arm-1 and Arm-2, respectively. 

Therefore, the velocity of the end-effector of Arm-1 can be rewritten as:
\begin{equation}
  \label{eq31}
\dot {r}_{e_i}=-(J_{b} H_{b}^{-1} H^1_{r}-J^1_{q}) \dot{\Theta}_{1}-J_{b} H_{b}^{-1} H^2_r \dot{\Theta}_{2}
\end{equation}

As we can see, the sophisticated coupling relationships between base and two manipulators makes kinematic equation contain many dynamic parameters hard to be identified. Meanwhile, the current state of Arm-1 depends on not only the its last joint trajectories, but also the joint trajectories in Arm-2 and vice versa. Thus, different from the trajectory planning of robots on the ground, trajectory planning of a free-float space manipulator is a complex time-varying problem. 

In addition, the dynamic model of a free-float dual-arm space manipulator can be derived as:
\begin{equation}
  \label{eq32}
\mathrm{H_s} \times \begin{bmatrix} {\ddot{\Theta}_{1}} \\{\ddot{\Theta}_{2}} \end{bmatrix} +\mathrm{C_s} \times \begin{bmatrix} {\dot{\Theta}_{1}} \\{\dot{\Theta}_{2}} \end{bmatrix} +\mathrm{C_b} \times \dot {r_{b}}=\begin{bmatrix} \mathbf \tau_1 \\\mathbf \tau_2 \end{bmatrix}
\end{equation}
where the detailed illustration of  $\mathrm{H_s}$ , $\mathrm{C_s}$ and $\mathrm{C_b} $ can be found in \cite{yan2021trajectory}.

\begin{figure*}[t]
  \centering
  \includegraphics[width=\hsize]{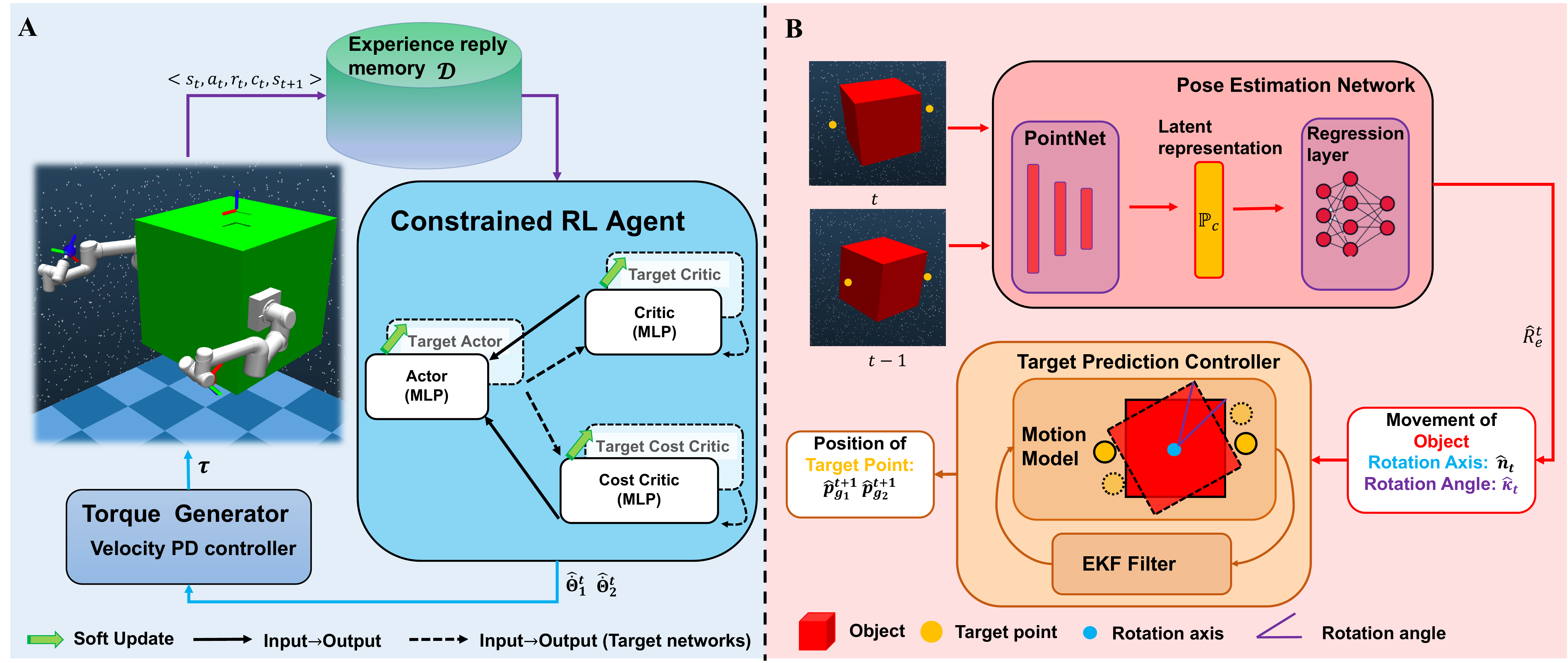}
  \caption{	\textbf{A}: Framework of Module I: constrained multi-target trajectory planning via reinforcement learning. \textbf{B}: Framework of Module II consists of pose estimation and target prediction. }
  \label{fig2}
\end{figure*}

\subsection{Constrained Markov Decision Process}

Before constructing the constrained reinforcement learning algorithm, we need to introduce the formulation of Constrained Markov Decision Process (CMDP), which contains $<S,A,P,\rho,R,C,\gamma_r,\gamma_c>$. Among them, $S$ is the state space, and $A$ is the action space. Additionally, $P:S\times A \times S \rightarrow \mathbb R$ represents the probabilistic transition function, and $\rho$ is the initial state distribution. $R: S \times A \rightarrow \mathbb R$ denotes the reward function with respect to the state and action, and $C:S \times A \rightarrow \mathbb R$ represents the cost function. Finally, $\gamma_r$ and $\gamma_c$ are the discount factor corresponding to reward and cost respectively, ranging from $[0,1)$. When training at time step $t$, the agent stays in a state $s_t$, then executes an action $a_t$, receives a reward $r_t$ and a cost $c_t$, and enters into a new state $s_{t+1}$. Notably, the action $a_t$ is generated by a policy $\pi(a_t|s_t)$, made by a deterministic policy $\mu(s_t)$ with a Gaussian noise in this paper. Besides, the total time steps in an episode $m$ is $T$, and the sum of episodes are $M$.

In order to maximize the total reward and satisfy the constraints, the objective of our problem is expressed as:
\begin{equation}
\label{eq4}
\begin{split}
& \max \ \mathcal J_r = {\mathbb E}_{s_0 \sim \rho,a_{0:T-1}\sim\pi,s_{1:T}\sim P} [\sum\nolimits_{t=0}^{T-1} \gamma_r^t r_t] \\ 
& \text { s.t. } \mathcal J_c = {\mathbb E}_{s_0 \sim \rho,a_{0:T-1}\sim\pi,s_{1:T}\sim P} [\sum\nolimits_{t=0}^{T-1} \gamma_c^t c_t]\leq C_s
\end{split}
\end{equation}
where $C_s$ represents the maximum threshold of the total cost. To estimate the relationships between reward, action and state, the state-action value and state-value functions have been constructed by:
\begin{equation}
  \label{eq5}
\begin{split}
Q^r_\pi(s,a)&={\mathbb E}_{s_t \sim P, a_t\sim \pi} [\sum\nolimits_{t=0}^{T-1} \gamma_r^t r_t| s_0= s,a_0=a] \\
V^r_\pi(s)&= {\mathbb E}_{a\sim \pi} [Q^r_\pi(s,a)]
\end{split}
\end{equation}
Similarly, we can define the state-action cost value and state cost value functions as $Q^c_\pi(s,a) $ and $V^c_\pi(s)$ respectively. The detailed derivation is omitted because of the same principle.

\section{Two-Module Motion Planning System}

% \subsection{System Overview}
The specific architecture of our two-module system is shown in Fig. \ref{fig2}. Aimed at tracking the moving target points on the object, the whole system consists of Module I and Module II. Firstly, Module I acts as an important tool for the trajectory planning of FFDASM for the targets within a large workspace. Importantly, the potential positions of targets are included in our workspace, thus making it possible to further track the targets. With the contribution of reinforcement learning, the constrained optimization problem we proposed can be solved successfully. Secondly, the core task of Module II is to estimate the pose of object, and then predict the position of targets on the object. According to the prediction, it can assist Module I to track the target points. As shown in Fig. \ref{fig2}, Point Cloud Net takes the point clouds of object between two frames as input, to estimate the rotation axis $\hat{n}_t$ and angle  $ \hat{\kappa}_t$.  Hereafter, we built the kinematic model of target points on the surface of object, and then applied Extended Kalman Filter to predict the positions of target points next step. Finally, Module I takes as input the prediction and other feedback to control the space manipulator. Different from the end-to-end model, our two-module system offers some advantages, like more scalable and interpretable learning process. It is also worthy noting that our study initially opens up new possibilities for implementing the combination of representation learning and reinforcement learning in the trajectory planning of FFDASM. In the following, we described the details of main parts of our two-module learning system. 

\section{Module I: Multi-Target Trajectory Planning Method}

\subsection{Formulation of Optimization Problem}

As discussed in the previous part, we transformed the trajectory planning problem into a reinforcement learning problem. To illustrate our optimization problem, we first introduced some concrete settings of components in MDP.

\textbf{State $S$}: the state vector $s_t$ at the step $t$ consists of the angular positions and velocities of joints, the positions of end-effectors and the position of target points of two manipulators, which means $s_t = [\Theta^t_1,\dot \Theta^t_1, \Theta^t_2, \dot\Theta^t_1, p^t_{e_1},p^t_{e_2},p^m_{g_1},p^m_{g_2}]$.  $p^m_{g_i}$ represents the target point of Arm-$i$, which is sampled in the work space $\mathcal W_i$  at the episode $m$. 

\textbf{Action $A$}: the action vector at the step $t$ is the desired velocities of joints, $a_t = [\hat{\dot {\Theta}}^t_1,\hat{\dot {\Theta}}^t_2]$.  Then a velocity tracking PD controller takes $a_t$ as input, to generate the torques of joints. Given that position controller performs less smoothly, it is better to choose the velocity controller \cite{Wang20221multi}.

\textbf{Probabilistic transition function $P$}: the core task of probabilistic transition function is to describe the stochastic transition  $s_{t+1} \rightarrow s_t$. As we expected, $P$ is corresponding to the dynamic model of our system, thereby resulting in the formulation: 
\begin{equation}
  \label{eq6}
\begin {aligned} s_{t+1} &\sim P (s_{t+1}|s_t,a_t) \\  &= \mathcal {F}(s_t,a_t,\delta_t) \end{aligned} \Leftrightarrow \begin{cases}
(3) \\
(4) \\
\mathbf \tau_t = K_{pd}(\hat{\dot {\Theta}}^t_1,\hat{\dot {\Theta}}^t_2) \\
\left\|*_{t}\right\| \leq U_{*} \ \ *\in [\Theta,\dot{\Theta},\boldsymbol{\tau}]\\
S_{SM} \cap \Omega =\varnothing
\end{cases}
\end{equation}
where $K_{pd}$ represents PD controller, $U_*$ is the maximum range of variables without the sake of safety, and $S_{SM}$ and $\Omega$ denote the space of space manipulators and base respectively. Additionally, $\delta_t$ is the uncertainty of our model.

\textbf{Reward  function $R$}: after executing an action, the agent will receive a reward based on the reward function $R(s_t,p^m_{g_{1:2}})$, where $p^m_{g_{1:2}}$ represents two target points at the episode $m$. In this paper, we utilized the sparse reward function, as Eq. \eqref{eq7} shows. As we can see, when the errors between end-effectors and target points are less than the thresholds $U_{e_{1}}$ and $U_{e_{2}}$, the reward is 0, else -1 at the step $t$. 
\begin{equation}
  \label{eq7}
    r_{t}=\{\begin{array}{c}
0, \text { if }\|p^{t}_{e_1}-p^m_{g_1}\| \leq U_{e_1}, \|p^{t}_{e_2}-p^m_{g_2}\| \leq U_{e_2} \\ 
-1, \text { else }
\end{array}  
\end{equation}

\textbf{Cost function $C$}: the prior studies \cite{2020Reinforcement,li2021constrained} didn't go further to explore the results under constraints. However, the violent fluctuation of base has a significant impact on communication of space manipulator, thereby leading to a failed task. In this case, we built a cost function with respect to the movement of base, as shown in Eq. \eqref{eq8}.
\begin{equation}
  \label{eq8}
c_t = \varkappa (r^t_{b}- r^0_{b})t
\end{equation}
Noticeably, if the final pose of base is closer to the initial pose, the sum of cost becomes fewer. 

Hereafter, following the above design, the trajectory planning of FFDASM can be defined as a reinforcement learning problem shown in Eq. \eqref{eq4}, maximizing the objective function $\mathcal J_r $ with the constraint $\mathcal J_c$.
\begin{equation}
  \label{eq9}
\begin{split}
\underset{\pi}{max} \ &\mathcal J_r = {\mathbb E}_{s_0,\delta} [\sum_{t=0}^{T-1} \gamma_r^t r_t]\\ 
\text { s.t. } &\mathcal J_c = {\mathbb E}_{s_0,\delta} [\sum_{t=0}^{T-1} \gamma_c^t c_t]\leq C_s \\
&s_{t+1} = \mathcal {F}(s_t,a_t,\delta_t)
\end{split}
\end{equation}

\subsection{Constrained Reinforcement Learning Algorithm}

Considering the constrained optimization problem and the requirement of multiple target points, we proposed a multi-target constrained reinforcement learning algorithm, Constrained Hindsight Experience Replay \textbf{CHER}. Firstly, the method constructs three networks including a state-action value network $Q^r_\pi(s_t,a_t, p^m_{g_{1:2}};\phi)$, a deterministic policy network $ \pi(\cdot|s_t,p^m_{g_{1:2}};\psi)$ and a cost-action value network $Q^r_\pi(s_t,a_t,p^m_{g_{1:2}};\eta)$. $\phi$, $\psi$  and $\eta$ are the parameters of networks. 

As we all know, the objective of  $Q^r_\pi(s_t,a_t,p^m_{g_{1:2}};\phi)$ and $Q^r_\pi(s_t,a_t,p^m_{g_{1:2}};\eta)$ is to estimate the sum of reward and cost with regard to the policy. Therefore, we can update $\phi$ by minimizing MSE loss of the TD error \cite{lillicrap2015continuous}: 
\begin{equation}
  \label{eq10}
\begin{split}
L_Q(\phi) &= E_{\mathcal D}[\frac{1}{2}(y_t-\nabla_{\phi_1}Q^r_\pi(s_t,a_t,p^m_{g_{1:2}};\phi))^2]\\
y_t &= r_t +\gamma_r \overline{Q}^r_\pi(s_{t+1},\overline{\pi}(\cdot|s_{t+1},p^m_{g_{1:2}};\psi^{\prime}),p^m_{g_{1:2}};\phi^{\prime})
\end{split}
\end{equation}
where $\mathcal D$ is the relay buffer including interaction information $<s_t,a_t,p^m_{g_{1:2}},r_t,s_{t+1}>$ at the step $t$ in the episode $m$,  $\overline{Q}^r_\pi$ is the target Q network parameterized by $\phi^\prime$, and $\overline{\pi}$ is the policy network parameterized by $\psi^\prime$. During the training, $\phi$ and $\psi$ are softly copied to $\phi^\prime$ and $\psi^\prime$ after some episodes repeatedly, which is shown as follows:

\begin{equation}
  \label{eq10-2}
\phi^\prime=\chi\phi^\prime+\left(1-\chi\right)\phi , \ \ \psi^\prime=\chi\psi^\prime+\left(1-\chi\right)\psi
\end{equation}
where $\chi \ll 1$, meaning that the update is slow and stable. The update of $Q^c_\pi(s_t,a_t,p^m_{g_{1:2}};\eta)$  is similar to Eq. \eqref{eq10}, so we omitted that for simplicity.

\textbf{Penalty method}: To address the problem, the first solution is the penalty method \cite{geibel2005risk}. Literally, the penalty method adds a penalty term with respect to the constraints into the previous objective, leading to a new objective function:
\begin{equation}
  \label{eq11}
\underset{\pi}{max} \mathcal J =\mathcal J_r - \lambda_p (\mathcal J_c - C_s)
\end{equation}
wherein $\lambda_p \geq 0$ is a constant depending on different tasks. Thanks to introducing $Q^r_\pi(s_t,a_t,p^m_{g_{1:2}};\phi)$ and $Q^r_\pi(s_t,a_t,p^m_{g_{1:2}};\eta)$, we can replace the sum of reward with the estimated Q function. That means the minimization objective of the policy network can be formulated as:
\begin{equation}
  \label{eq12}
L_\pi(\psi) = E_{\mathcal D}[-Q^r_\pi(s_t,a_t,p^m_{g_{1:2}};\phi)+\lambda_p (Q^c_\pi(s_t,a_t,p^m_{g_{1:2}};\eta)-C_s)]
\end{equation}
When $\lambda_p$ is larger, the policy will avoid the risk of increasing the cost more carefully.

\textbf{Lagrangian method}: other than the penalty method, the optimization problem can also be transformed into its Lagrangian dual form to optimize \cite{stooke2020responsive}. The optimal parameters of policy network is defined as:
\begin{equation}
  \label{eq13}
\psi^* = \underset{\psi}{\arg \max} \ \underset{\lambda_l \geq 0}{\min} \ E_{\mathcal D} [ Q^r_\pi(s_t,a_t,p^m_{g_{1:2}};\phi)-\lambda_l (Q^c_\pi(s_t,a_t,p^m_{g_{1:2}};\eta)-C_s)]
\end{equation}
where $\lambda_l$ is the Lagrangian multiplier that strike the balance between the constraint and objective. Similar to the penalty method, the update of policy network is approximated by according to stochastic gradient descent:

\begin{equation}
  \label{eq14}
\begin{split}
\nabla_{\psi} L_\pi(\psi) \propto E_{\mathcal D}[(-\nabla_{a_t} Q^r_{\pi}(s_t, a_t,p^m_{g_{1:2}};\phi) 
+ \lambda_l \nabla_{a_t} Q^c_{\pi}(s_t, a_t,p^m_{g_{1:2}};\eta))\nabla_{\psi} \pi(\cdot|s_t;\psi) ]    
\end{split}
\end{equation}
Furthermore, during training, the Lagrangian multiplier $\lambda_l$ is updated by the following objective:
\begin{equation}
  \label{eq15}
\begin{split}
L(\lambda_l) = \lambda_l (Q^c_\pi(s_t,a_t;\eta)-C_s) \\
\lambda_l \leftarrow \max (0, \lambda_l + \zeta \nabla_{\lambda_l} L(\lambda_l) )   
\end{split}
\end{equation}
where $\lambda_l$ is the learning rate of Lagrangian multiplier. 

\textbf{Hindsight Experience Replay (HER)}: Moreover, to mitigate the sparsity of reward in multi-goal problem, we utilized the Hindsight Experience Replay (HER) method to improve the sample efficiency \cite{andrychowicz2017hindsight}. To be specific, the method assumes that the positions of two target points $p^m_{g_1}$ and $p^m_{g_2}$ at the episode $m$ are the previous goals, and then the final positions of end-effectors $p^{T}_{e_1}$ and $p^{T}_{e_2}$ replace them as new goals. That means $[p^{T}_{e_1},p^{T}_{e_2}] \Rightarrow [\hat p^m_{g_1},\hat p^m_{g_2}]$. Meanwhile, the reward will be recalculated based on new goals, thus converting a failed experience into a successful experience. 

To recap, after $M$ episodes, the optimal parameters of three networks $Q^r_\pi(s_t,a_t,p^m_{g_{1:2}};\phi)$, $Q^c_\pi(s_t,a_t,p^m_{g_{1:2}};\eta)$ and $ \pi(\cdot|s_t,p^m_{g_{1:2}};\psi)$ can be provided by the above minimization objectives.  Algorithm \ref{alg:1} depicts the pseudo code for training CHER. Among them, $\pi_b(\cdot|s_t,p^m_{g_{1:2}};\psi)$ denotes a noisy version of current policy for exploration during the training.

\begin{algorithm}[t]
	\renewcommand{\algorithmicrequire}{\textbf{Input:}}
	\renewcommand{\algorithmicensure}{\textbf{Output:}}
	\caption{ Constrained Hindsight Experience Replay (CHER)}
	\label{alg:1}
	\begin{algorithmic}[1]
	    \STATE Orthogonal initialize $\phi,\phi^{\prime},\psi,\psi^{\prime}, \eta,\eta^{\prime},\lambda_p, \lambda_l$
	    \STATE Initialize replay buffer $\mathcal D$
	    \FOR {episode $m = 1, M $}
	     \STATE Sample targets $p^m_{g_{1}}, p^m_{g_{2}}$ and initial state $s_0$ 
	    \FOR {step $t=0, T-1$}
	    \STATE Sample a action from $\pi_b(\cdot|s_t,p^m_{g_{1:2}};\psi)$
	    
	    \STATE Execute the action $a_t$ and observe a new state $s_{t+1}$
	    \ENDFOR
	    \FOR {step $t=0, T-1$}
	    \STATE $r_t:=R(s_t,p^m_{g_{1:2}})$
	    \STATE Store $<s_t,a_t,p^m_{g_{1:2}},r_t,s_{t+1}>$ into $\mathcal{D}$
	    \STATE $ [\hat p^m_{g_1},\hat p^m_{g_2}] := [p^{T}_{e_1},p^{T}_{e_2}]$
	    \STATE $r^\prime_t:=R(s_t,\hat p^m_{g_{1:2}})$
	    \STATE Store $<s_t,a_t,\hat p^m_{g_{1:2}},r^\prime_t,s_{t+1}>$ into $\mathcal{D}$
	    \ENDFOR
	    \FOR{iteration $n = 1, N$}
	    \STATE Sample a minibatch $\mathcal B$ from the replay buffer $\mathcal D$
	    \IF{Penalty method}
	    \STATE Update the $\psi,\psi^{\prime},\phi,\phi^{\prime}$ via Eq.\eqref{eq10},\eqref{eq10-2},\eqref{eq12} using minibatch $\mathcal B$
	    \ENDIF
	    \IF{Lagrangian method}
	    \STATE Update the $\psi,\psi^{\prime},\phi,\phi^{\prime},\lambda_l$ via Eq.\eqref{eq10},\eqref{eq10-2},\eqref{eq14},\eqref{eq15} using minibatch $\mathcal B$
	    \ENDIF
	    \ENDFOR
	    \ENDFOR
	\end{algorithmic}
\end{algorithm}

\section{Module II: Pose Estimation and Prediction Method }

\subsection{Pose Estimation based on Point Clouds}

The essential function of Module II is to construct an integrated system that can estimate the pose of the object and then predict the position of target points for trajectory planning. Notably, under the intense influence of sunlight in space, it is difficult for space manipulators to estimate the pose via RGBD images. Therefore, we just consider the point clouds of objects obtained by LiDAR sensors, which are more accurate in real-world practices. 

To be specific, an intuitive approach is to take the point cloud at this frame as input, and the estimated pose of the object as output. However, the results extremely depend on the global coordinate frame. Thus, it is beneficial to utilize the point clouds between two frames to evaluate the relative rotation matrix. Referring to \cite{huang2021generalization}, we also built a 3-layered PointNet \cite{qi2017pointnet} parameterized by $\omega$, to encode two point clouds. Then, we extracted the latent representations $ \mathbb {P}_c$ by the MLP neural network. Additionally, the batch normalization layer is added behind each fully connected layer to accelerate the training. Finally, the network outputs the predicted pose vector $\vec{r_e}$, and rewrite it as $\hat{R_e}$. Following \cite{suwajanakorn2018discovery}, the training objective can be expressed as:
\begin{equation}
  \label{eq16}
\underset{\omega} {min} \ L_{R} = 2 \arcsin (\frac {1}{2\sqrt{2}}\|R_p-\hat{R_e}(\mathbb P_c;\omega)\|_F)
\end{equation}
where $R_p$ is the ground truth of the rotation matrix. In the training process, we sampled random points on the surface of objects, and then constructed two point clouds by randomly rotating the original point cloud. Thus, the label is the true relative rotation between them, as well as the input of networks is two generated point clouds. 

To sum up, the first part of Module II achieves the relative pose estimation between two frames. Although the predicted rotation matrix still can not directly provide the prior information for the latter target prediction, we found that the predicted rotation can facilitate the recovery of the relative rotation axis $\hat{n}_t$ and angle $ \hat{\kappa}_t$ of object at the step $t$ based on Eq. \eqref{eq17}.
\begin{equation}
  \label{eq17}
\hat {\kappa}_t =\operatorname{\arccos}\left(\frac{\operatorname{tr}(\hat {R_e^t})-1}{2}\right) ,\ \hat {R_e^t} \hat{n}_t = \hat{n}_t
\end{equation}

Furthermore, it is worth noting that though the traditional ICP \cite{pomerleau2015review} method can recover the rotation axis and angle by an iterative optimization, the critical advantage of a learning system is higher scalability and generalization. The latent representation $\mathbb P_c $ provided by PointNet can benefit the classification and detection of objects after, which is discussed in the experiments part.

\subsection{Target Prediction via Extended Kalman Filter}

According to the estimated relative rotation axis $\hat{n}_t$ and angle  $ \hat{\kappa}_t$ of the object at the step $t$, we can construct the kinematic model of the target point on the surface of the object. Firstly, without the effect of external forces, we assume the object moves in a regular circle around the rotation axis $n_r$, which means $\hat{n}_t$ is close to a constant. In practice, this assumption is easily satisfied in a low-gravity environment \cite{li2021constrained}. Secondly, the target point is attached to the surface of object, thereby with the same rotating velocity. 

In the context of the above assumptions, we built the motion and observation models for target point as shown in Eq. \eqref{eq18}, where the state vector $ X^i_{t}$ represents $[x^i_t, y^i_t,\varsigma^i_t,v^i_t] $ observation vector $Z^i_t$ represents $[\widetilde{x}^i_t, \widetilde{y}^i_t]$, and $\delta_t$ is the control cycle. Among them, $x^i_t$ and $y^i_t$ are the position of the target point in the rotating plane $\mathcal S_{\perp n_r}$, $\varsigma^i_t$ is the rotation angle, and $v^i_t$ is the velocity of the target point. Notably, $x^i_t$ and $y^i_t$ can be derived by the projection transformation $\mathbb T (p^t_{g_i};\mathcal S_{\perp n_r})$, where $p_{t_i}$ is the position of the target point of Arm-$i$. To indicate the gap between the model and reality, some uncertain terms are included in two models. Specifically, $\varpi^i_t$ and $\varrho^i_t$ denote the process and measurement noises.
\begin{equation}
  \label{eq18}
\begin{split}
\begin{cases}
x^i_{t+1} = x^i_t+v^i_t cos(\varsigma^i_t) \delta_t \\
y^i_{t+1} = y^i_t+v^i_t sin(\varsigma^i_t) \delta_t \\
\varsigma^i_{t+1} = \varsigma^i_t + \dot{\hat{\kappa}}^i_t  \delta_t \\
v^i_{t+1} = v^i_t
\
\end{cases}
\Rightarrow X^i_{t+1} = f(X^i_t) + \varpi^i_t, \ \varpi^i_t \sim N(\textbf 0, \mathcal Q^i)  
\end{split}
\end{equation}

\begin{equation}
  \label{eq19}
Z^i_{t+1} = [\textbf{I}_{2\times2}\ \textbf{0}] X^i_t + \varrho^i_t = h (X^i_t) + \varrho^i_t, \ \varrho^i_t \sim N(\textbf 0, \mathcal R^i)
\end{equation}

Since the models are non-linear, we can use the Extended Kalman Filter (EKF) \cite{fujii2013extended} to obtain the predicted state vector $\hat X^i_{t+1}$, as Eq. \eqref{eq20} shows.
\begin{equation}
  \label{eq20}
\hat X^i_{t+1} = f(X^i_t) + \textbf K^i_{t} (Z^i_t - h(f(X^i_t)))
\end{equation}
$\textbf K^i_t$ represents the Kalman gain derived by EKF method at the step $t$, and mathematical details can be found in \cite{fujii2013extended}. Following the above equation, we have the access to predict the position of the target point in real time. Based on the previous projection transformation, the position of the target point for Arm-$i$ can be written as:
\begin{equation}
  \label{eq21}
\hat p_{g_i}^{t+1} =\mathbb T (\hat {x}^i_{t+1}, \hat {y}^i_{t+1};\mathcal S_{\perp n_r})
\end{equation}

Remarkably, when $\hat p_{g_i}^{t+1}$ is taken as an input in Module I at the step $t$, the strategy can reach the dynamic target point within a bounded time and error. We proposed an upper bound of convergent time.

\textbf{Theorem 1}: The velocity of target point is less than the velocity of end-effector, which means $v_t \leq \dot p^t_{e_i}, t \in T $, and the error between $\hat p_{g_i}^{t+1}$ and $p_{g_i}^{t+1}$ is bounded by $\epsilon_{p_{g_i}} $. Thus, we can guarantee the formulation satisfied, when the initial error is $ d_{e_i}$: 
\begin{equation}
  \label{eq22}
\| p^{T_B}_{e_i}-p^{T_B}_{g_i} \| \leq U_{e_i}+\epsilon_{p_{g_i}} ,  d_{e_i} \leq \int^{T_B}_{t} (\dot p^t_{e_i} - v_t) \ dt
\end{equation}
The minimum of $T_B$ is the upper bound of convergent time, thereby meaning the end-effector will track the target point after $T_B$.

\section{EXPERIMENTS}

\subsection{Comparison with Other Baselines}

\begin{figure*}[!t]
  \centering
  \includegraphics[width=\hsize]{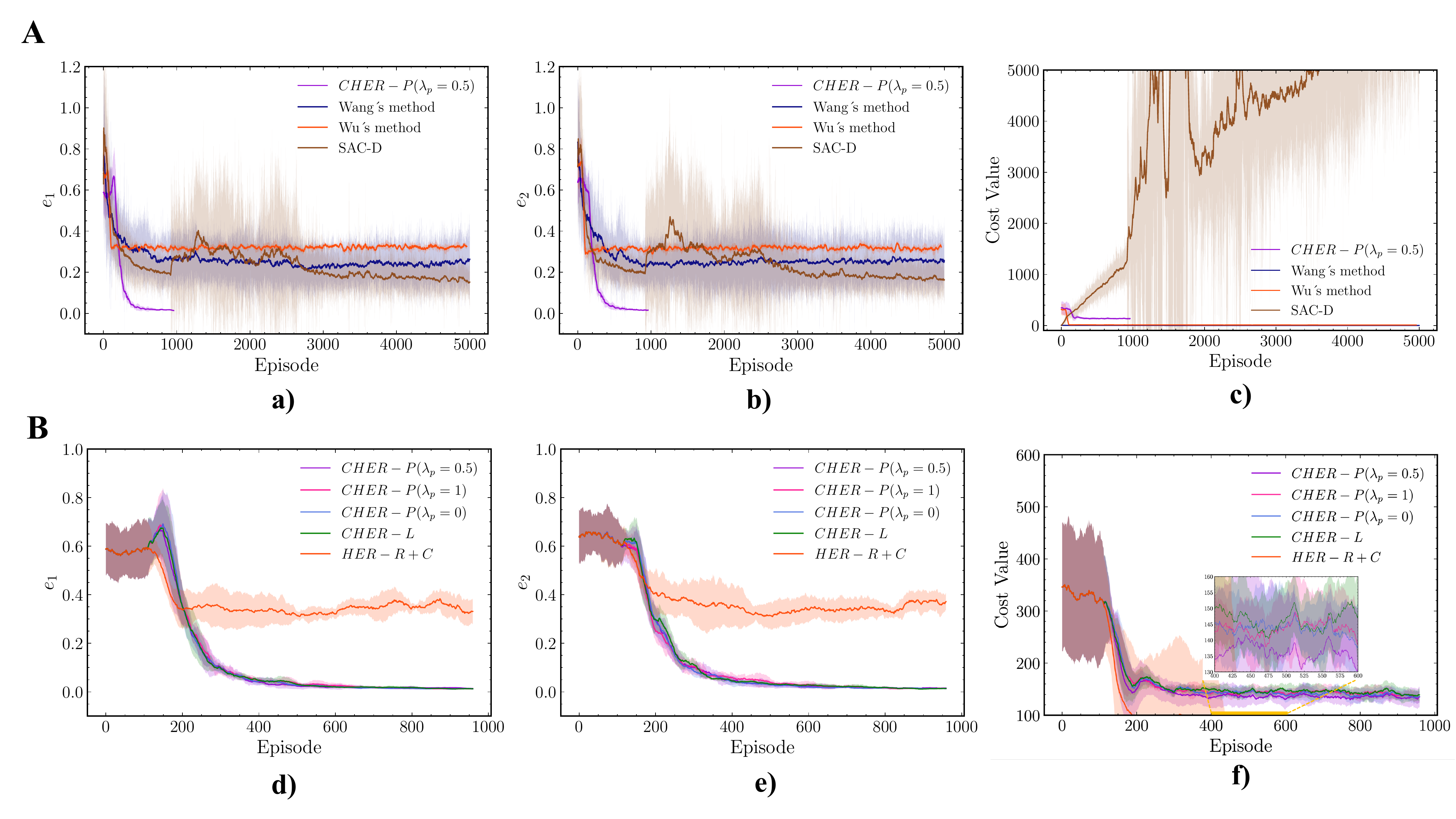}
  \caption{\textbf{A}: a-c) Position errors and cost value using our method and baseline methods. \textbf{B}: d-f) Position errors and cost value based on different constrained setting. (Curves smoothed report the mean of 5 random seeds, and shadow areas show the variance.)}
  \label{fig3}
\end{figure*}

The goal of Module I is to provide a trajectory planning method for multiple targets in a large work space. We chose 3 state-of-the-art RL-based methods for FFDASM as other baselines. 
\begin{itemize}
    \item \textbf{Wu's method \cite{2020Reinforcement}}: Wu's method realized the trajectory planning for 8-Dof free-float dual-arm space manipulators for the single target.
    \item \textbf{Wang's method \cite{Wang20221multi}}: Wang's method solved the task of multiple targets for a single arm based on an improved version of PPO algorithm.
    \item \textbf{SAC-D}: Considering Soft Actor-Critic (SAC) algorithm achieves superior performance in many robotic control task \cite{haarnoja2018soft}, we designed SAC-D algorithm for our task, in which we applied the similar reward function in \cite{Wang20221multi}.
\end{itemize}
Compared with other baselines for FFDASM, the experimental results of our method illustrate a significant improvement in planning accuracy. Concretely, as shown in Fig. \ref{fig3}-A, our method \textbf{CHER-P} (Penalty method $\lambda_p =0.5$) facilitates the end-effector to achieve the task within the least number of episodes. Admittedly, other algorithms have performed well in some tasks. However, multi-target trajectory planning for 12-DoF dual-arm space manipulators remains a critical challenge for most algorithms, as shown in Fig. \ref{fig3}-A. This is because the increasing dimensions of state and action have a huge impact on the efficiency of exploration, especially for multiple targets. Our method (CHER-P($\lambda_p =0.5$)) introduced HER method to improve sample efficiency, facilitating to speed up the convergence. Note that because other algorithms is unable to satisfy the basic need of planning accuracy, we overlook the requirements of constraints in their training. 

Furthermore, to evaluate the performance of accuracy and constraints together, we plotted $e_1$, $e_2$ and cost value in testing episodes for 3 algorithms, which are as follows:
\begin{itemize}
    \item \textbf{CHER-P} (Penalty method) : $\lambda_p$ is a constant coefficient of constraints . 
    \item \textbf{CHER-L} (Lagrangian method): $\lambda_l$ is a varying coefficient of constraints, which can be updated by Eq. \eqref{eq15} iteratively.
    \item \textbf{HER-R+C}: the method based Hindsight Experience Replay (HER) models the trajectory planning as a Markov Decision Process (MDP) instead of CMDP. Thus, A new reward function can be derived as $R_{\text{new}}= R+C$, where $R$ and $C$ are above-described.
\end{itemize}
As Fig. \ref{fig3}-B shows, it can be observed that the algorithms based HER except for 
\textbf{HER-R+C} achieve comparable performances on planning accuracy. That means if the reward function contains multiple types of objectives (rewards or costs), it is hard to strike a balance between these demands during training. Additionally, from the perspective of the cost value, although \textbf{CHER-L} can adjust the Lagrangian multiplier $\lambda_l$ according to the violation of constraints dynamically, the final stage in the training will exist significant constraint violations. Alternatively, in this paper, \textbf{CHER-P} ($\lambda_p =0.5$) strikes the most suitable balance between reward and cost. To recap, the penalty coefficient of constraints plays an important role in practical usage. To sum up, Table \ref{tab1} clearly illustrates learned policy performance for different reinforcement learning algorithms, and we can see that the versions of our method have superior results in contrast to others.

Finally, It is worthy noting that all algorithms used the same model to built networks, which consists of four fully connected layers with ReLU activation. The number of hidden units per layer is 256, and the activation of output layer in policy network is Tanh function. Additionally, the input and output are also the same variables, and other parameters in each algorithm are provided by some papers \cite{2020Reinforcement,Wang20221multi} or OpenAI baselines \cite{baselines}.

\begin{table}[!htb]
  \centering
  \caption{The best performance of different learning algorithms.The best results are marked as bold. Note that we considered its cost value is valid only if an algorithm meets the demand, $e_{1}$ or $e_{2} \leq 0.05 $.}
  \label{tab1}
  \resizebox{.55\columnwidth}{!}{
      \begin{tabular}{l|l|l|l}
        \hline
       Algorithm   & $e_1$ (m) & $e_2$ (m)   & Cost Value \\ \hline
        \textbf{Wu's method \cite{2020Reinforcement}} & $0.324 \pm 0.012$ & $0.323 \pm 0.010$ &  $9.765 \pm 0.839$ \\ \hline
        \textbf{Wang's method \cite{Wang20221multi}} & $0.255 \pm 0.110$ & $0.171 \pm 0.085$ & $8.196 \pm 0.384$ \\ \hline
        \textbf{SAC-D} & $0.096 \pm 0.077$ & $0.100 \pm 0.031$ & $5844 \pm 1846$ \\ \hline
        \textbf{HER-R+C} & $0.331 \pm 0.052$ & $0.371 \pm 0.035$ & $40.462 \pm 8.620$ \\ \hline
        \textbf{CHER-L} & $0.013 \pm 0.002$ & $\pmb{0.014 \pm 0.002}$ & $139.207 \pm 12.170$ \\ \hline
        \textbf{CHER-P($\lambda_p=0$)} & $\pmb{0.012 \pm 0.002}$ & $\pmb{0.014 \pm 0.002}$ & $135.440 \pm 11.918$ \\ \hline
        \textbf{CHER-P($\lambda_p=1$)} & $\pmb{0.012 \pm 0.002}$ & $0.014 \pm 0.003$ & $134.865 \pm 11.404$ \\ \hline
        \textbf{CHER-P($\lambda_p=0.5$)} & $0.014 \pm 0.002$ & $0.015 \pm 0.004$ & $\pmb{131.838 \pm 10.558}$ \\ \hline
      \end{tabular}
  }
\end{table}

\subsection{Robustness of Trajectory Planning}

Utilizing the model trained by \textbf{CHER-P} algorithm, we evaluated the performance of trajectory planning under random five pairs of starts and ends in working space. Most importantly, the testing environment is different from that of the training, because the initial position of end-effector is fixed in training. In this case, 100 repeated results shown in Fig. \ref{fig3}-C demonstrate our method has a strong robustness. What we should also highlight is the smoothness of trajectory our method generated. We can observe that two end-effectors reach the targets quickly and converge to a narrow region near targets. Although the final positions of the two end-effectors meet the demand of the threshold we set, the final error has a slight difference from each other. Interestingly, we found the same situation in many previous studies \cite{2020Reinforcement,li2021constrained}. Intuitively, the reason behind that is the coupling effect between two arms make it hard to balance two individual reaching tasks for them. Also, we guessed multi-agent learning will amend the bias, which will be discussed in future work.

In addition, considering the gap between the real world and simulation, we tested our trained policy under different scenarios where the mass of base is changing. Fig. \ref{fig4} shows the robust property of our method clearly. Even though the mass of the base decreases by nearly 50\%, the planning error doesn't saw a significant increase finally. This is because such a sufficient exploration in multi-target space reduces the sensitivity of disturbance. The above results demonstrate our method have obtained considerable robustness and generalization abilities, so as to have the potential to transfer the model to the real world.

\subsection{Generalization of Pose Estimation}

\begin{figure}[t]
  \centering
  \includegraphics[width=\hsize]{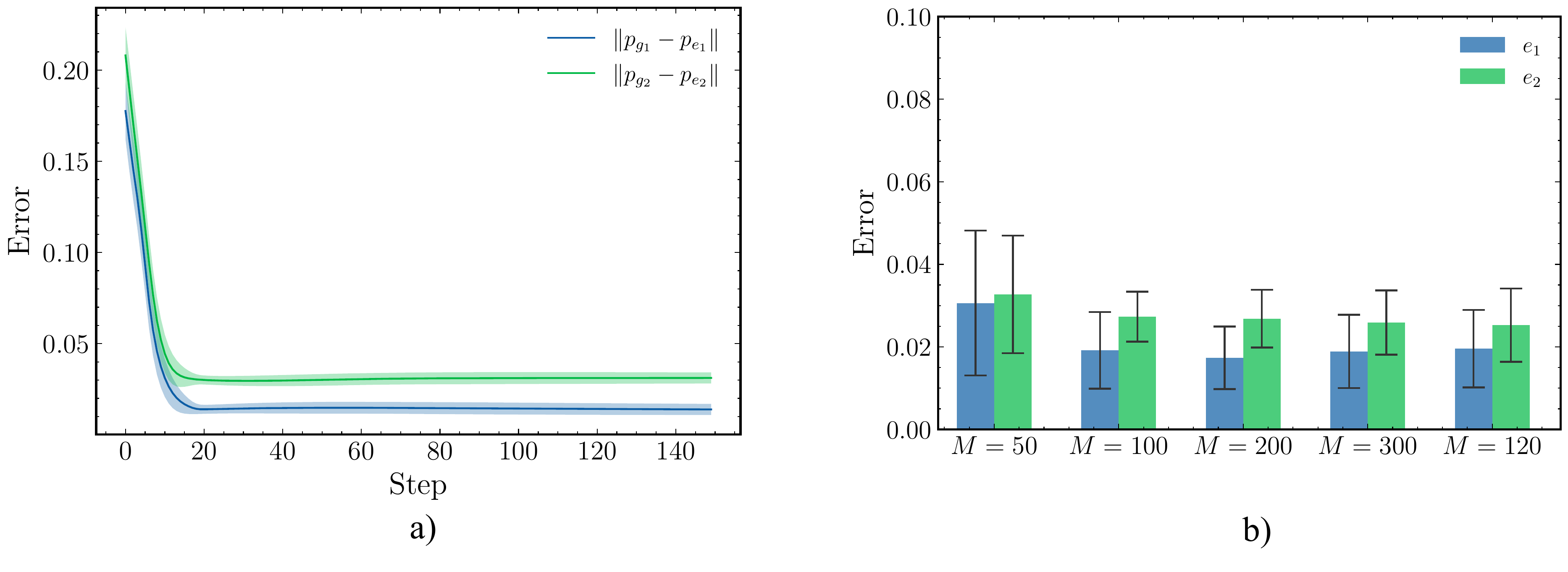}
  \caption{a): Errors $e_1$ and $e_2$ during an episode under different initial position of end-effectors. b): Errors $e_1$ and $e_2$ under different masses of base.}
  \label{fig4}
\end{figure}

In the first part of Module II, the pose estimation network based on point clouds of the object recovers the rotation axis and angle of object in space. In order to enhance the generalization, we didn't make full use of point clouds, but sampled 128 points from the surface of the object, adding some Gaussian noises to the sampled points. Meanwhile, we introduced their normal vector to facilitate the convergence rate. Furthermore, although the shape of object mainly is a cube, we extended the types of object, adding two open-source object datasets, ContactDB \cite{brahmbhatt2019contactdb} and YCB \cite{calli2015ycb}. Actually, a better way is to use specific objects in space, but there is no related sources, which will be our future direction. Because the scale of objects are not consistent, we normalized the point clouds of objects before training. Fig. \ref{fig4}-a illustrates the pose errors in the testing environment, where the cubic object (a non-cooperative object) rotates in 5 different speeds. We can see that the relative errors remain at the lowest point under the highest spinning speed. This is because the relative difference of two point clouds is slight in two frames, leading to the ambiguity of latent representation $\mathbb {P}_c$.  As a matter of fact, in real-world practices, we run the process of pose estimation in a low frequency to solve the problem, because the object's movement is sustained for a long term in space. Furthermore, to better exploit $ \mathbb {P}_c$, we took $ \mathbb {P}_c$ as input for the classification task. Based on the MLP network and cross-entropy loss, the testing accuracy is $93.34\pm 5.3\%$, illustrating our framework is scalable and effective. Finally, we appreciate the released code in \cite{huang2021generalization}, because we designed our experiments referring to it. 

\begin{figure}[t]
  \centering
  \includegraphics[width=\hsize]{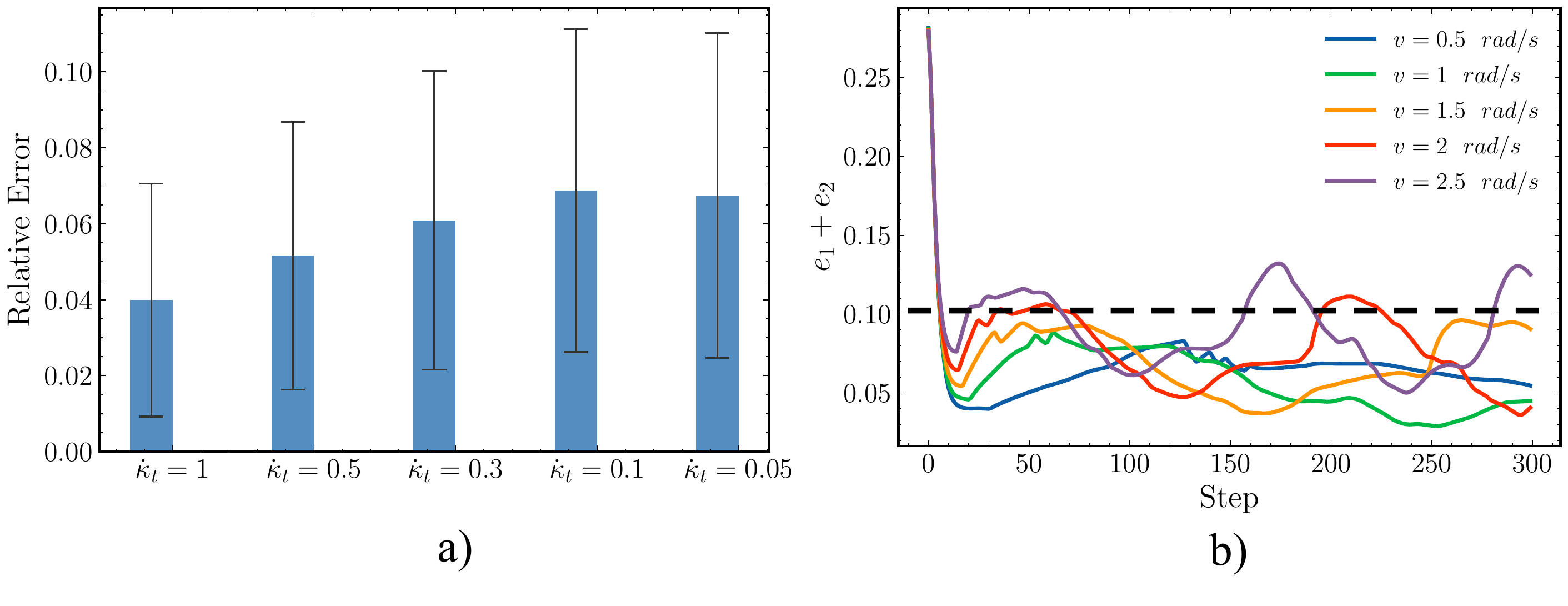}
  \caption{a): Relationship between relative error and spinning speed of object. b): Curves of the sum of errors $e_1+e_2$ under different spinning speeds.}
  \label{fig5}
\end{figure}

\begin{figure}[!t]
  \centering
  \includegraphics[width=\hsize]{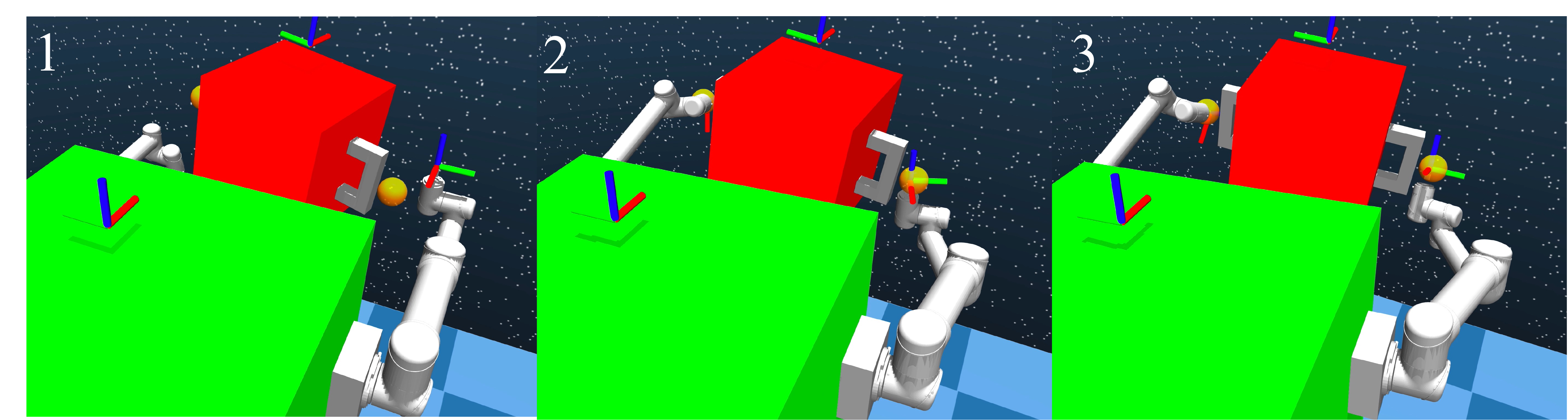}
  \caption{Video snapshots of the process that space manipulator tracks the target points.}
  \label{fig6}
\end{figure}

After estimating the relative pose of the object between two frames, we can obtain the rotation axis and angle of the object during the period according to Eq. \eqref{eq17}. Then, according to our motion model of target points, we can get the predicted position of target points based on EKF method next step. 
% In this setup, we assumed the positions of target points come from the prior information before, thus we added some noises on the real positions. During the experiment, we just considered the prediction process of one target points on a spinning object, but the changing of these situations will have no impact on the performance of our method. As we can see, the predicted trajectory (red line) has the slight difference with the real trajectory of target point (blue line). Therefore, when we employ this technique multiple times, we will obtain the possible position of targets at the next step in real time. 

\section{Validation of Target Tracking}
After estimating the relative pose of the object between two frames, we can obtain the rotation axis and angle of object during the period according to Eq. \eqref{eq17}. Then, according to our motion model of target points, we can get the predicted position of target points based on EKF method next step. 
With the access to predicted target positions, our method can realize tracking dynamic targets in space. Firstly, we set a testing environment, where the spinning speed of targets is ranging from $0.5 \ rad/s$ to $2.5 \ rad/s$, and the radius of rotation is $0.15 \ m$. As shown in Fig. \ref{fig5}, the end-effector will track the target point successfully. However, we can notice that when the spinning speed is $2.5 \ rad/s$, the sum of errors $e_1+e_2$ shows an upward trend that means a sign of un-convergence. The underlying reason can be analyzed by Theorem 1. Specifically, when the velocities of end-effectors are less than those of targets, convergent time $T_B$ is $\infty$ and the sum of errors can not be bounded by $U_{e_i}+\epsilon_{p_{g_i}}$. 

Furthermore, we simulated a practical scenario to better exploit the advantage of our system. After an occurrence of a failure, a target satellite uncontrollably rotates around an axis. The commander wants a free-float dual-arm space manipulator to catch the target satellite. After several orbital transfers, our manipulator stays a certain distance with the target satellite. Then, LiDAR sensors on space manipulator receives the point clouds of target satellite, and the pose estimation network take them as input to estimate the rotation axis and angular speed. Importantly, based on kinematic characteristics of target satellite, target prediction controller can predict the future positions of target points for end-effectors. Finally, because the trajectory of target points is involved in our training work space, Module I allows the end-effectors to track the moving target points. Fig. \ref{fig6} shows the video snapshots of the process. While in this work we didn't consider the behavior capturing the target satellite, it can be further extended, which we leave as an avenue for future work.

\section{CONCLUSIONS}

We proposed a learning-based motion planning system for a free-float dual-arm space manipulator to track dynamic targets, which contains Module I and Module II. For Module I, the comparison experiments demonstrate our multi-target trajectory planning algorithm is able to reach the targets within a large workspace for a 12-DoF dual-arm space manipulator. Meanwhile, taking advantage of Penalty method and Lagrangian method, our algorithm strikes a balance between objective and constraints efficiently. In this case, our study facilitates the RL-based method to be applied in space with multiple sophisticated constraints. Moreover, the combination of representation learning and EKF method allows Module II to estimate the pose of object and predict the position of target points. Importantly, the extended experiments illustrate our method has high scalability and generalization so as to achieve a promising result on a practical scenario of tracking targets on a non-cooperative object.

\section*{Acknowledgments}
This work was supported in part by the National Natural Science Foundation of China under Grant U21B6002. 

%Bibliography
\bibliographystyle{unsrt}  
\bibliography{templateArxiv}

\end{document}